\documentclass[conference]{IEEEtran}
\IEEEoverridecommandlockouts
\usepackage{cite}
\usepackage{amsmath,amssymb,amsfonts}
\usepackage{algorithmic}
\usepackage{graphicx}
\usepackage{textcomp}
\usepackage{xcolor}
\usepackage{caption} 
\usepackage{multirow}
\usepackage{booktabs}
\usepackage{arydshln}
\usepackage{newfloat}
\usepackage{listings}
\usepackage{algorithm}
\usepackage{algorithmic}
\def\BibTeX{{\rm B\kern-.05em{\sc i\kern-.025em b}\kern-.08em
    T\kern-.1667em\lower.7ex\hbox{E}\kern-.125emX}}
\begin{document}

\title{HARIS: Human-Like Attention for Reference Image Segmentation

\thanks{$^*$: Corresponding Author}
}

\author{\IEEEauthorblockN{1\textsuperscript{st} Mengxi Zhang}
\IEEEauthorblockA{\textit{School of Electrical and Information Engineering} \\
\textit{Tianjin University}\\
Tianjin, China. \\
mengxizhang@tju.edu.cn}
\and
\IEEEauthorblockN{2\textsuperscript{nd} Heqing Lian}
\IEEEauthorblockA{\textit{Xiao Ying AI Lab} \\
\textit{Xiao Ying Company}\\
Beijing, China \\
lianheqing@xiaoyingai.com}
\and
\IEEEauthorblockN{3\textsuperscript{rd} Yiming Liu}
\IEEEauthorblockA{\textit{Xiao Ying AI Lab} \\
\textit{Xiao Ying Company}\\
Beijing, China \\
liuyiming@xiaoyingai.com}
\and
\IEEEauthorblockN{5\textsuperscript{th} Jie Chen$^*$}
\IEEEauthorblockA{\textit{School of Electronics Engineering and Computer Science} \\
\textit{Peking University}\\
Shenzhen, China \\
chenj@pcl.ac.cn}
}

\maketitle

\begin{abstract}
Referring image segmentation (RIS) aims to locate the particular region corresponding to the language expression. Existing methods incorporate features from different modalities in a \emph{bottom-up} manner. This design may get some unnecessary image-text pairs, which leads to an inaccurate segmentation mask. In this paper, we propose a referring image segmentation method called HARIS, which introduces the Human-Like Attention mechanism and uses the parameter-efficient fine-tuning (PEFT) framework. To be specific, the Human-Like Attention gets a \emph{feedback} signal from multi-modal features, which makes the network center on the specific objects and discard the irrelevant image-text pairs. Besides, we introduce the PEFT framework to preserve the zero-shot ability of pre-trained encoders. Extensive experiments on three widely used RIS benchmarks and the PhraseCut dataset demonstrate that our method achieves state-of-the-art performance and great zero-shot ability.
\end{abstract}


\section{Introduction}
Referring image segmentation~\cite{hu2016segmentation} (RIS) is a typical multi-modal task, which aims to locate the particular region corresponding to the language expression. Different from conventional image segmentation that generates masks of some fixed categories, the target of RIS is to locate the referents according to the free-form language expressions. Therefore, the main challenge for RIS is to get a comprehensive vision-language understanding and effectively fuse features from different modalities.

\begin{figure}[t]
	\flushleft{\includegraphics[scale=0.25]{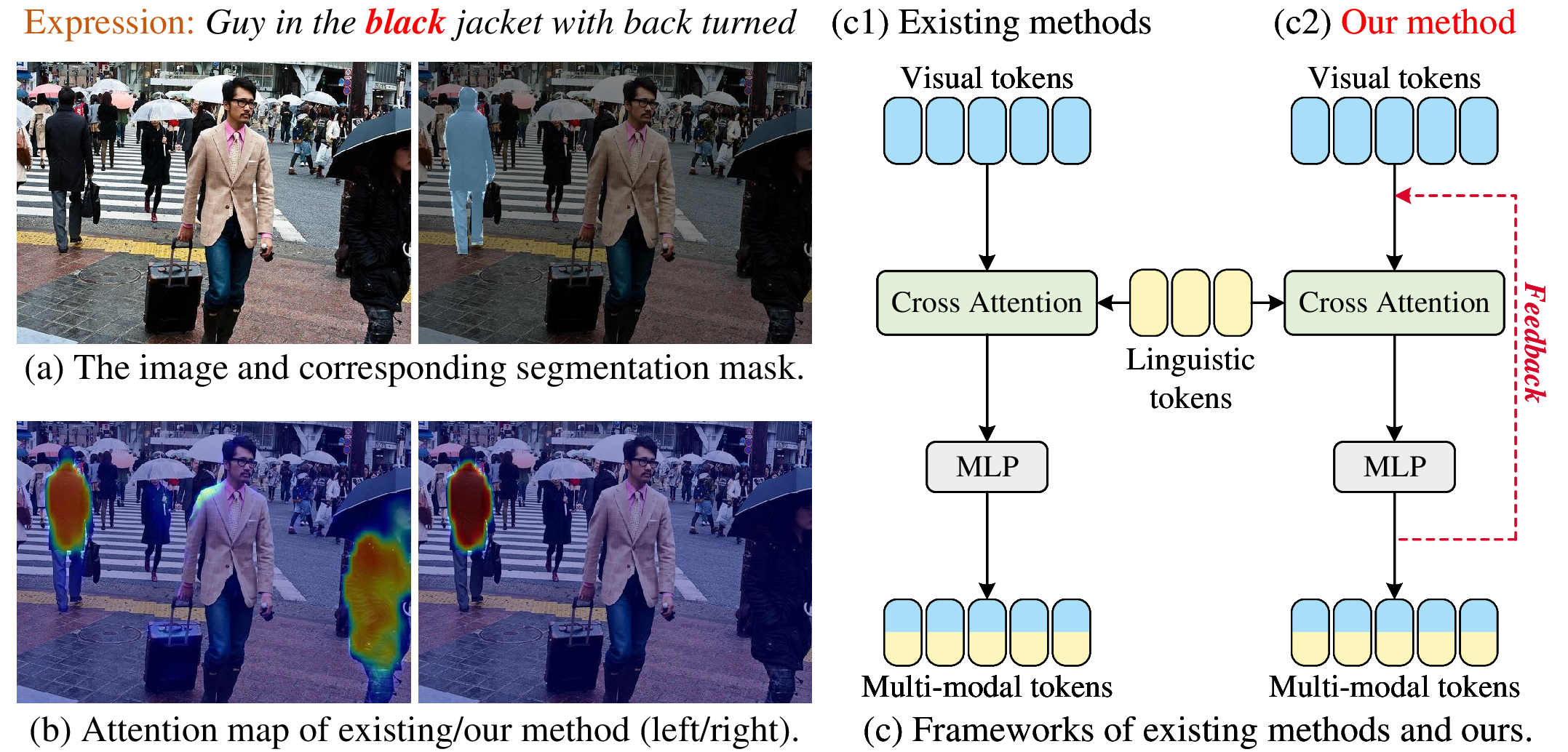}}
	\caption{The attention maps and frameworks of existing methods and our method. Existing methods fuse features from different modalities based on the cross-attention mechanism, which follows a \emph{bottom-up} manner, as shown in the left part of (c). As a result, the attention map generated by these methods may contain some irrelevant image-text pairs (the left part of (b)). For example, the correct region for \emph{`black'} is the left part of the image. However, existing methods also take the right part as relevant regions of \emph{`black'}. Different from existing methods, our method (the right part of (c)) introduces a feedback signal, which comes from modulated multi-modal tokens. Therefore, our method gets an accurate region for the word \emph{`black'} (the right part of (b)).}
 \label{framework}
\end{figure}

Existing RIS methods design different methods to fuse visual and linguistic features. Recently, some methods~\cite{cris, vlt, gres} introduce the attention mechanism for vision-language fusion and achieve significant improvement. CRIS~\cite{cris} adopts the cross-attention mechanism to get multi-modal features. ReLA~\cite{gres} designs the region-based attention to explicitly model relationships between image regions and each word. However, these methods only explore the vision-language relationships in a \emph{bottom-up} manner, which may lead to some unnecessary vision-language pairs. As shown in Fig.~\ref{framework}, only the guy on the left accords with the expression \emph{`Guy in the black jacket with his back turned'}. However, conventional fusion methods take both the right and left image regions as the relevant objects for \emph{`black'} due to the bottom-up mechanism. 

\begin{figure*}[t]
	\centering{\includegraphics[scale=0.28]{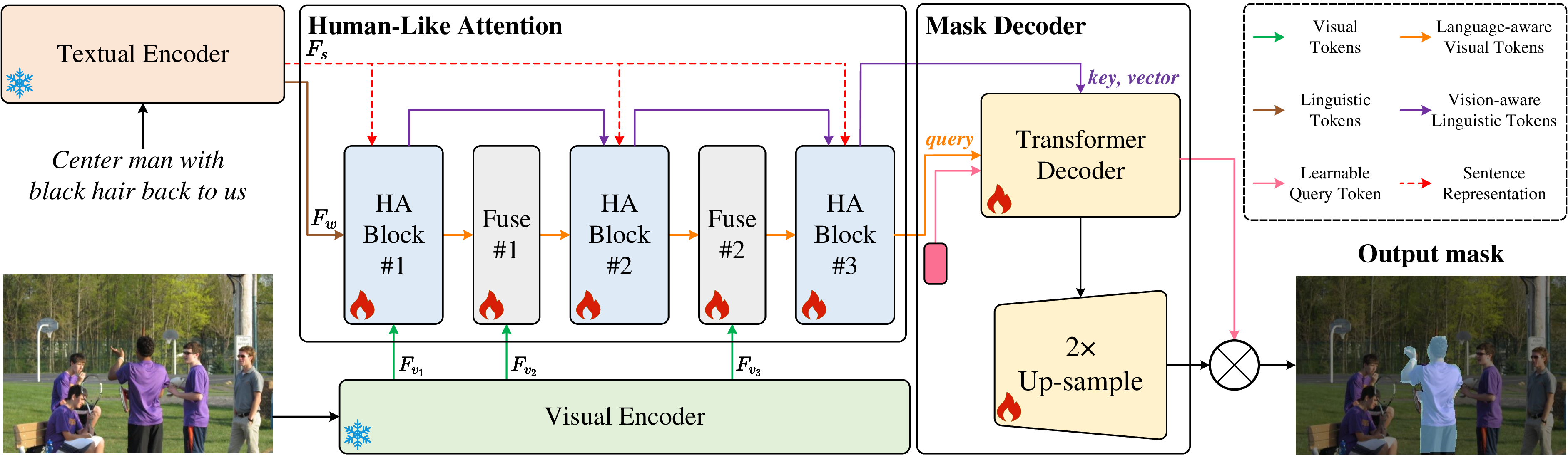}}
	\caption{
 The overview of HARIS. The input image is fed into the image encoder and outputs visual features ($F_{v_1},F_{v_2},F_{v_3}$) from different layers. Correspondingly, we send the language expression to the text encoder and get linguistic features $F_w$ and sentence representation $F_s$. Then, these features are sent into the Human-Like Attention blocks to get multi-modal features. Besides, we use a hierarchical architecture to use both semantic and grained visual features. Last, we send the multi-modal features and a learnable query token to the Transformer Decoder and get the final segmentation mask.}
    \label{overview}
\end{figure*}

In this paper, we propose a novel RIS method that uses the Human-Like Attention mechanism based on the parameter-efficient fine-tuning framework. First, we design the Human-Like Attention to avoid unnecessary vision-language pairs.
Unlike conventional bottom-up fusion methods, we introduce an additional \emph{feedback} signal derived from multi-modal features. This signal is integrated with the visual tokens, serving as the secondary input for the attention block. This innovative design is inspired by bionics. Specifically, in human's vision and thought, the cognition of things usually iteratively advances, using new information to refine existing knowledge. Our Human-Like Attention mechanism pushes the network to revisit feature fusion across varied modalities.  Thus, our method makes the network center on the specific object and neglects the irrelevant ones, as depicted in Fig.~\ref{framework}, where the right part about \emph{`black'} is discarded.
Second, we leverage the parameter-efficient fine-tuning (PEFT) framework to preserve the zero-shot ability of the visual and textual encoder. In the training process, we only set the parameters of the fusion block and mask decoder learnable, which saves computational resources and avoids catastrophic forgetting. To be summarized, our main contributions are summarized as follows:

$\bullet$ We design a novel RIS method called HARIS, which leverages the PEFT framework to preserve the great generation ability for encoders of different modalities.

$\bullet$ We propose Human-Like Attention to reduce the interference from unnecessary vision-language pairs, which makes the network focus on the referred object.

$\bullet$ Our approach exhibits state-of-the-art performance on three widely used RIS datasets, $i.e.$, RefCOCO, RefCOCO+, and G-Ref. Additionally, our method achieves excellent zero-shot ability on the PhraseCut dataset.

\section{Related Work}
\subsection{Referring Image Segmentation}
Prior RIS approaches~\cite{yu2018mattnet,liu2019learning} concatenate visual and linguistic features before sending them to the convolution neural networks (CNN) to generate multi-modal features. However, these methods are sub-optimal for multi-modal fusion due to the limitations of CNN modeling. Since the attention mechanism demonstrates promising performance in diverse domains, certain methods~\cite{vlt, gres, kim2022restr} exploit the attention mechanism in the field of RIS. ReSTR~\cite{kim2022restr} leverages the Transformer architecture to capture long-range dependencies of different modalities. More recently,  ReLA~\cite{gres} proposes the region-based attention mechanism to model the region-region and region-language dependencies explicitly. However, these methods only use \emph{bottom-up attention}, which may lead to some irrelevant vision-language pairs. To address this issue, we propose the Human-Like Attention mechanism to make the model rethink the relationships between visual and linguistic features, similar to human cognitive processes.

\subsection{Attention Mechanism}
Recently, the attention mechanism has achieved great success in various vision-language tasks. Pioneer methods~\cite{blip,blipv2} use the Transformer-decoder-based architecture to align visual and linguistic features. Later, GLIP~\cite{glip} introduces a bidirectional attention mechanism to align these features more precisely. Recently, the multi-modal large language models (MLLM) have demonstrated great performance on various multi-modal tasks. Typically, the architectures of MLLM are a stack of Transformers. In this paper, we exploit the practicality of top-down attention in RIS, which may inspire other vision-language tasks.

\begin{figure}[t]
	\centering{\includegraphics[scale=0.26]{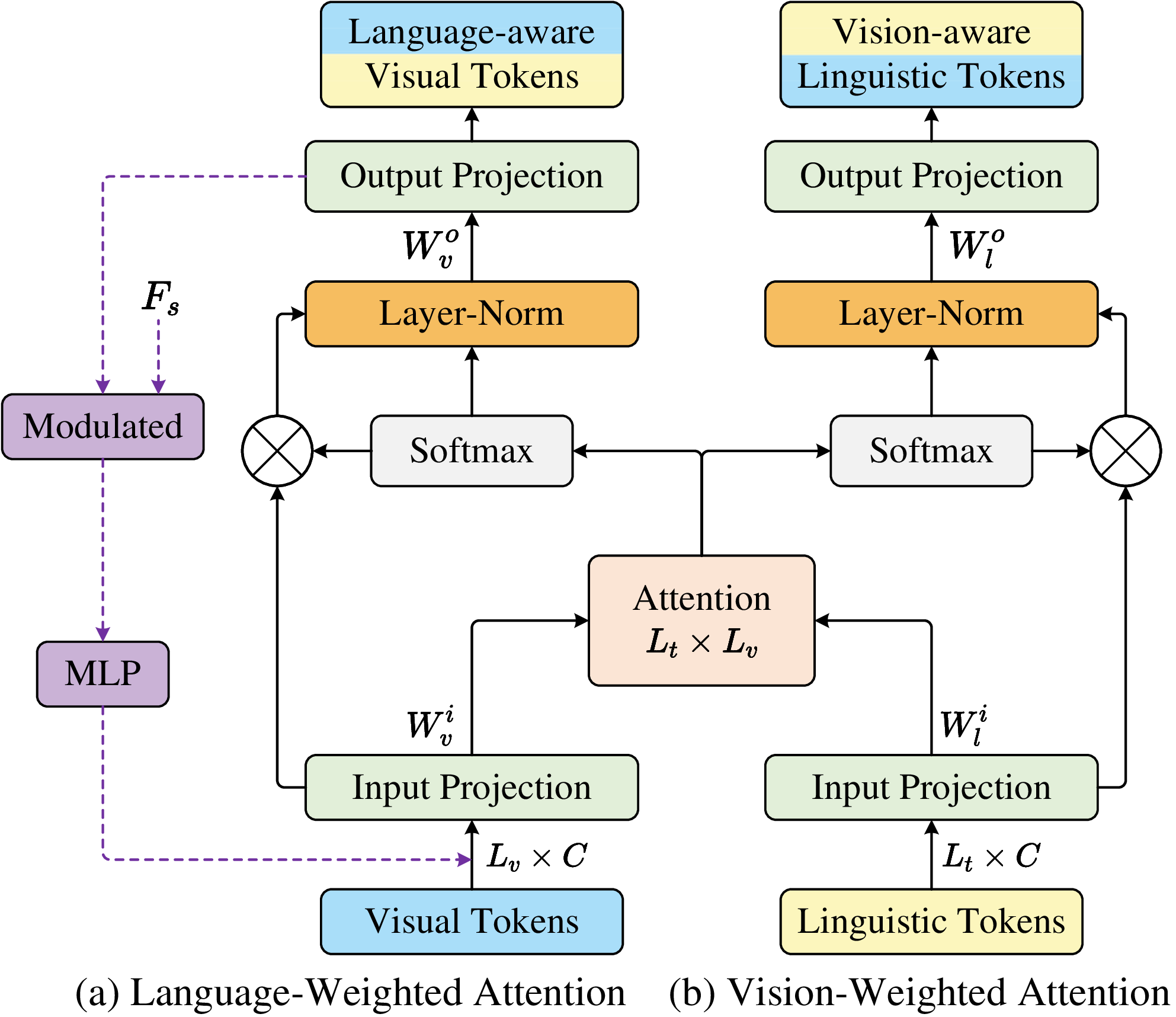}}
	\caption{The architecture of Human-Like Attention block. This block consists of two branches: Language-Weighted Attention and Vision-Weighted Attention. In the Language-Weighted Attention, we introduce the feedback signal, which is modulated by $F_s$ and fed into the MLP layer. Then, the feedback signal together with visual tokens acts as the second-round inputs for Language-Weighted Attention.}
 \label{ha}
\end{figure}

\section{Method}
The overall framework of HARIS is shown in Fig.~\ref{overview}. We first extract visual features ($F_{v_1}, F_{v_2}, F_{v_3}$) and linguistic features ($F_{w}, F_{s}$) from frozen image encoder and text encoder, respectively. Specifically, $F_{v_1}/F_{v_2}/F_{v_3} \in \mathbb{R}^{H\times W\times C}$ symbolize the visual features from shallow/middle/deep layers, where $H$ and $W$ denote height and width of the feature map, $C$ is the number of channels. For the convenience of feature fusion, we flatten these visual features into a sequence, forming them into the shape of $L_v\times C$, $L_v=H\times W$. $F_{w}\in \mathbb{R}^{L_t\times C_t}$ denotes feature for each word, and $F_{s}\in \mathbb{R}^{1\times C_t}$ is the representation of the whole sentence. The sentence length is denoted as $L_t$, while the channel number of linguistic features is represented by $C_t$.

Then, we send the visual features and linguistic features into the Human-Like Attention block to obtain the multi-modal features. In particular, we utilize both semantic and grained information by a hierarchical architecture. Finally, the Mask Decoder takes the multi-modal features to get the segmentation mask. The details of each module are described in the following section.

\subsection{Human-Like Attention Block}
Previous methods fuse visual features based on the bottom-up attention mechanism. Although these methods show satisfactory results, they may produce some irrelevant image-text pairs, as shown in Fig.~\ref{framework}. To address this issue, we propose the integration of Human-Like Attention block, leveraging an additional feedback signal to eliminate redundant image-text pairs. The architecture of the Human-Like Attention block is depicted in Fig.~\ref{ha}.

First, we model the relationships between visual tokens and linguistic tokens as follows,

\begin{equation}
    \begin{split}
         &E_v = F_{v}W_{v}^{i}, E_l = F_{l}W_{l}^{i}, \\
         &A = \mathrm{Softmax}(\frac{E_v E^{\top}_l}{\sqrt{C}}),
    \end{split}
\label{HA:1}
\end{equation}
where $W_{v}^{i} \in C\times C$ and $W_{l}^{i}\in C_t\times C$ are two learnable matrices to transform tokens of $F_{v}$ and $F_{l}$ from different modalities into the same feature dimension. $A\in L_v \times L_t$ is the attention matrix. $\frac{1}{\sqrt{C}}$ denotes the scale factor.

Then we get the multi-modal features in a bidirectional way, which is formulated by,

\begin{equation}
    \begin{split}
         F_{l2v} = &\mathrm{LayerNorm}(AE_{l} + E_{v})W_{v}^{o},  \\
         F_{v2l} = &\mathrm{LayerNorm}(A^{\top}E_{v} + E_{l})W_{l}^{o},
    \end{split}
\label{HA:2}
\end{equation}
where $\mathrm{LayerNorm}(\cdot)$ denotes the layer-normalization layer. $W_{v}^{o}$ and $W_{l}^{o}$ denote weights of linear layers for mapping. $F_{l2v}$ and $F_{v2l}$ symbolize language-aware visual tokens and vision-aware linguistic tokens, which are the outputs of Language-Weighted Attention and Vision-Weighted Attention, respectively. After we get these multi-modal features, we design an extra branch to get the feedback signal. This branch can be viewed as the human's thinking process, where they use new knowledge to refine existing knowledge. Such a design makes the model center on the referring objects accurately. The feedback branch is shown by the purple dash lines in Fig.~\ref{ha}. 

Specifically, we first use the whole sentence representation $F_{s}^{\top}$ to get modulated features $\overline{F}_{l2v}$. The mathematical process is shown below.

\begin{equation}
     \overline{F}_{l2v} = \mathrm{Softmax}(\frac{F_{l2v} F^{\top}_s}{\sqrt{C}})F_s + F_{l2v}.
\label{HA:3}
\end{equation}

Then, the feedback signal is obtained by a Multi-Layer Perceptron (MLP). We add the feedback signal to the visual tokens as the second-round inputs of Language-Weighted Attention. Finally, we run the feed-forward path of Language-Weighted Attention again and get the final language-aware visual tokens.

\begin{table*}[t]
\setlength{\tabcolsep}{7pt} 
\centering
\caption{IoU comparisons with previous state-of-the-art methods. U: UMD split; G: Google split.}
\label{tab:comp} 
\begin{tabular}{c|ccc|ccc|ccc}
\hline
\multicolumn{1}{c|}{\multirow{2}{*}{Methods}} & \multicolumn{3}{c|}{RefCOCO} & \multicolumn{3}{c|}{RefCOCO+} & \multicolumn{3}{c}{G-Ref}    \\ \cline{2-10} 
\multicolumn{1}{l|}{}   & val      & testA   & testB   & val      & testA    & testB   & val (U) & test (U) & val (G) \\ \hline
ReSTR (CVPR 2022)      & 67.22             & 69.30               & 64.45               & 55.78    & 60.44    & 48.27   & -      & -       & 54.48   \\
CRIS (CVPR 2022)       & 70.47             & 73.18               & 66.10               & 62.27    & 68.08    & 53.68   & 59.87   & 60.36    & \_      \\
ETRIS (ICCV 2023)       & 70.51             & 73.51               & 66.63               & 60.10    & 66.89    & 50.17   & 59.82   & 59.91    & 57.88      \\
RefTR (NIPS 2021)      & 70.56             & 73.49               & 66.57               & 61.08    & 64.69    & 52.73   & 58.73   & 58.51    & \_      \\
LAVT (CVPR 2022)       & 72.73             & 75.82               & 68.79               & 62.14    & 68.38    & 55.10   & 61.24   & 62.09    & 60.50   \\
VLT (TPAMI 2022)       & 72.96             & 75.96               & 69.60               & 63.53    & 68.43    & \underline{56.92}   & 63.49   & \underline{66.22}    & \underline{62.80}   \\
MCRES (CVPR 2023)     & \underline{74.92} & \underline{76.98}   & \underline{70.84}   & \underline{64.32}      & \underline{69.68}             & 56.64           & \underline{63.51}         & 64.90             & 61.63   \\ \hline
\textbf{HARIS (Ours)}                         & \textbf{76.17}    & \textbf{78.45}      & \textbf{71.37}      & \textbf{66.01}    & \textbf{71.92}    & \textbf{57.02}   & \textbf{65.05}   & \textbf{66.60}    & \textbf{64.55}    \\ \hline

\end{tabular}
\end{table*}

\subsection{Hierarchical Design}
A high-quality referring image segmentation mask requires both global and local information, which serves for precise location and accurate details. To this end, we introduce the hierarchical design to utilize both semantic and grained visual features: ${F}_{v1}$,${F}_{v2}$, and ${F}_{v3}$. Specifically, we fuse visual features from different layers after each Human-Like Attention block. Take $\mathrm{Fuse}$ \#1 as the example, the mathematical process is formulated as follows,

\begin{equation}
     \widehat{F}_{v_2} = \mathrm{CBA}(\mathrm{Concat}(\overline{F}_{l2v}, F_{v_2})),
\label{HD:1}
\end{equation}
where $\mathrm{CBA}(\cdot)$ is the sequential operation consisting of convolutional layer with the kernel size of $3\times 3$ and stride $1$. $\widehat{F}_{v_2}$ is the input for next Human-Like Attention Block \#2.

\subsection{Mask Decoder}
To fully utilize the multi-modal tokens from dual attention branches, we build the Mask Decoder based on the standard Transformer Decoder architecture~\cite{detr, mask2former}. In particular, we concatenate a learnable query token $M$ with vision-aware linguistic tokens as the query of the Transformer. Correspondingly, we set the language-aware visual tokens as the key and value. In the training process, the learnable query token assembles the linguistic information and interacts with the visual information, thus getting a comprehensive vision-language understanding. This process is formulated as follows.

\begin{equation}
     F^{m} = \mathrm{TD}(\mathrm{Concat}(M, F_{l2v}), F_{v2l}, F_{v2l}),
\label{HA:3}
\end{equation}
where $\mathrm{TD}(\mathrm{q,k,v})$ represents a standard Transformer Decoder with the input of query/key/vector ($\mathrm{q/k/v}$). $M$ symbolizes the learnable query token. $\mathrm{Concat}(\cdot,\cdot)$ represents the concatenation operation.

After that, we up-sample $F^{m}$ by two sequential blocks, each of which contains a convolution layer, a batch-normalization layer, and an up-sample operation. Finally, we multiply the up-sampled features with the learnable query token $M$ and get the final segmentation mask.

\subsection{Loss}
We adopt the linear combination of focal loss~\cite{lin2017focal} $\mathcal{L}_f$ and dice loss~\cite{dice} $\mathcal{L}_d$ as the optimizing target. The overall loss $\mathcal{L}$  is calculated as follows.
\begin{equation}
     \mathcal{L} = \mathcal{L}_f + \mathcal{L}_d.
\label{loss}
\end{equation}

\section{Experiments}
\subsection{Datasets and Metrics}
We evaluate our method on three typical RIS datasets, $i.e.$, RefCOCO \& RefCOCO+~\cite{refcoco}, and G-Ref~\cite{GRef}. These three datasets, sourced from MSCOCO~\cite{coco}, are annotated with different linguistic styles. The average expression length of RefCOCO/RefCOCO+ is approximately 3.61/3.53 words, respectively. Notably, RefCOCO+ differs from RefCOCO by excluding expressions related to absolute positions, such as 'left/right'. G-Ref has a longer average expression length of 8.4 words. Similar to prior studies, we evaluate RefCOCO and RefCOCO+ across validation/testA/testB splits. For G-Ref, we employ both partitions of UMD and Google.

As for the evaluation metrics, we adopt two widely used metrics, $i.e.$, mask Intersection-over-Union (IoU) score and precision with thresholds (Pr@$X$). The IoU score reflects the quality of predicted masks, while Pr@$X$ is the proportion of predicted masks that achieve an IoU score exceeding a specific threshold $X\in \{70,80,90\}$. 

\subsection{Implementation Details}
We utilize a visual encoder pre-trained by~\cite{sam} and a text encoder pre-trained by ~\cite{clip}. 
During the training process, we set the parameters of these encoders frozen. The maximum word length is tailored to each dataset: for RefCOCO and RefCOCO+, it is set at 17 words; while for G-Ref, the limit is extended to 25 words. We train the network with the Adam optimizer for 50 epochs, where the initial learning rate is $1\textrm{e}^{-4}$ and decreases by a factor of 0.1 at the 30th epoch. All experiments are conducted on 4 Nvidia V100 with a batch size of 16.

\subsection{Main Results}
We compare our proposed method, HARIS, with a series of previous state-of-the-art (SOTA) methods on three widely-used datasets, $i.e.$, RefCOCO, RefCOCO+, and G-Ref. According to Tab.~\ref{tab:comp}, our approach exceeds others on each split of all datasets even with frozen visual and text encoders.

On the RefCOCO dataset, the IoU performance of our proposed method surpasses the second-best SOTA method, $i.e.$, MCRES~\cite{mcres}. To be specific, our method achieves $1.25\%$, $1.47\%$, and $0.53\%$ IoU gain on the val, testA, and testB split, respectively. The significant improvement reveals the effectiveness of our proposed HARIS. 

Besides, our HARIS also achieves new SOTA performance by $1.69\%/2.24\%/0.10\%$ on three splits, which verifies that our method is also competitive for expressions without absolute positioning, such as \emph{`left'}, \emph{`right'}.

Finally, on the most challenging G-Ref dataset where the length and style are various, our method improves over other SOTA methods by $1.54\%$, $0.38\%$, and $1.75\%$ IoU on the val (U), test (U), and val (G) split, respectively. Based on these significant improvements, we assert that our method gets a more holistic image-text understanding ability and thus gets a high-quality segmentation mask.

\subsection{Ablation Study}
We conduct ablation experiments to investigate the specific contributions of each component of HARIS on the RefCOCO val split.. The first experiment (\# 1) removes the hierarchical structure (HS) and only utilizes the features from the last layer of the visual encoder. In the second experiment (\# 2), the decoder is replaced by point-wise multiplying multi-modal features and globally averaged language features, which is similar to previous works~\cite{cris}. In the third experiment (\# 3), we replace our Human-Like Attention Block with a conventional cross-attention block. 

\subsubsection{Effectiveness of Human-Like Attention}
To alleviate irrelevant image-text pairs, we design the Human-Like Attention mechanism. As illustrated in Tab.~\ref{tab:abl_vla}, the replacement of the conventional cross-attention (CA) block brings $1.45\%$ descent (\# 3), which indicates the effectiveness of the feedback signal. To get a more comprehensive understanding of Human-Like Attention, we conduct a series of experiments to evaluate each design in Human-Like Attention. First, we remove the feedback (FB) branch and only preserve the bidirectional attention mechanism. 
As shown in Tab.~\ref{tab:abl_vla}, this strategy (\# 5) results in $1.11\%$ IoU drop. This is because the feedback branch reduces some unnecessary image-text pairs. Then, we explore the impacts of Vision-Weighted (VW) Attention (\# 6) and Language-Weighted (LW) Attention (\# 7). According to Tab.~\ref{tab:abl_vla}, Vision-Weighted (VW) Attention and Language-Weighted (LW) Attention bring $1.28\%$ and $3.31\%$ IoU improvement, respectively. These improvements indicate that Language-Weighted Attention is more important than Vision-Weighted Attention.

\begingroup
\setlength{\tabcolsep}{5.pt} 
\renewcommand{\arraystretch}{1.15} 
\begin{table}[t]
\centering
\caption{The ablation study of each component in our HARIS.}
\label{tab:abl_vla} 
\begin{tabular}{c|ccccc}
\hline
Settings & Methods & Pr@70 & Pr@80 & Pr@90 & IoU   \\ \hline
\# 1     & w/o HS   & 74.87 & 65.42  & 38.16 & 71.75 \\
\# 2     & r. DE & 79.26 & 69.98 & 41.15  & 74.98 \\     
\# 3     & r. CA    & 77.45 & 67.67 & 41.08 & 74.72 \\\hline 
\# 4     & \textbf{Ours}    & \textbf{80.35} & \textbf{71.79} & \textbf{42.72} & \textbf{76.17}  \\ \hline

\end{tabular}
\end{table}

\begingroup
\setlength{\tabcolsep}{5.pt} 
\renewcommand{\arraystretch}{1.15} 
\begin{table}[t]
\centering
\caption{The ablation study of each component in our HARIS.}
\label{tab:ha} 
\begin{tabular}{c|ccccc}
\hline
Settings & Methods & Pr@70 & Pr@80 & Pr@90 & IoU   \\ \hline
\# 5     & w/o FB  & 77.41 & 68.74 & 40.62 & 75.06 \\
\# 6     & w/o VW  & 78.02 & 68.43 & 41.36 & 74.89 \\ 
\# 7     & w/o LW  & 75.96 & 65.95 & 38.74 & 72.86 \\ \hline
\# 4     & \textbf{Ours}    & \textbf{80.35} & \textbf{71.79} & \textbf{42.72} & \textbf{76.17}  \\ \hline

\end{tabular}
\end{table}

\subsubsection{Effectiveness of PEFT Framework}
To demonstrate the superiority of the PEFT framework, we compare our method with others in terms of zero-shot ability. Specifically, we use the test split of PhraseCut ~\cite{wu2020phrasecut} to emphasize the zero-shot feature. PhraseCut contains 1287 categories, much more abundant than the 80 categories in COCO. We report the zero-shot performance of our method and others in Tab.~\ref{tab:zero-shot}. 

\begin{table}[ht]
\setlength{\tabcolsep}{13.5pt} 
\centering
\caption{Zero-shot performance of different methods.}
\label{tab:zero-shot}
\begin{tabular}{c|ccc}
\hline
\multirow{2}{*}{Training Set} & \multicolumn{3}{c}{IoU results on PhraseCut} \\ \cline{2-4} 
                              & CRIS  & LAVT  & \textbf{Ours}   \\ \hline
RefCOCO                       & 15.53 & 16.68 & \textbf{21.62} \\
RefCOCO+                      & 16.30 & 16.64 & \textbf{21.30} \\
G-Ref                         & 16.24 & 16.05 & \textbf{22.93} \\ \hline
\end{tabular}
\end{table}

As illustrated in Tab.~\ref{tab:zero-shot}, the zero-shot ability of our method significantly exceeds previous methods. This property benefits from the PEFT framework, which preserves the zero-shot ability of these encoders. In contrast, other methods set these encoders trainable to get a great performance on specific datasets. However, the number of these specific datasets is insufficient for the convergence of encoders and thus destroys the zero-shot ability.

\section{Conclusion}
In this paper, we propose a novel referring image segmentation method called HARIS, which introduces the Human-Like Attention mechanism based on the parameter-efficient fine-tuning framework. Specifically, we introduce an extra feedback branch from multi-modal features and feed it into the attention block as the second-round input. This design alleviates some irrelevant image-text pairs. Besides, we introduce the parameter-efficient fine-tuning framework into RIS, which preserves the zero-shot ability of the pre-trained encoders. Extensive experiments on three widely used benchmarks and PhraseCut demonstrate that HARIS achieves new state-of-the-art performance and great zero-shot ability.

\bibliographystyle{IEEEbib}
\bibliography{icme}

\end{document}